\documentclass[a4paper,conference]{IEEEtran}
\ifCLASSINFOpdf
\else
\fi
\usepackage{acronym}
\newacro{CNN}[CNN]{Convolutional Neural Network}
\newacro{BoCF}[BoCF]{Bag of Color Features}
\newacro{FC4}[FC$^4$]{Fully Convolutional Color Constancy}
\newacro{MCDE}[MCDE]{Monte Carlo Dropout Ensembles}

\usepackage{graphicx}
\usepackage{amsmath,amssymb} 

\begin{document}
%
\title{Monte Carlo Dropout Ensembles for Robust Illumination Estimation}

\author{\IEEEauthorblockN{Firas Laakom\IEEEauthorrefmark{1},
Jenni Raitoharju\IEEEauthorrefmark{2},  Alexandros Iosifidis\IEEEauthorrefmark{3}, Jarno Nikkanen\IEEEauthorrefmark{4}, and Moncef Gabbouj\IEEEauthorrefmark{1}}\\
\IEEEauthorblockA{\IEEEauthorrefmark{1}Department of Computing Sciences and Communication Sciences, Tampere University, Finland\\
\IEEEauthorrefmark{2}Programme for Environmental Information, Finnish Environment Institute, Finland\\
\IEEEauthorrefmark{3}Department of Engineering, Aarhus University, Denmark\\
\IEEEauthorrefmark{4}XIAOMI, Hermiankatu 6-8H,  Tampere, Finland\\
Emails: \IEEEauthorrefmark{1}firas.laakom@tuni.fi,
\IEEEauthorrefmark{2}jenni.raitoharju@environment.fi,
\IEEEauthorrefmark{3}alexandros.iosifidis@eng.au.dk, \\
\IEEEauthorrefmark{4}jarnon@xiaomi.com,
\IEEEauthorrefmark{1}moncef.gabbouj@tuni.fi}}


\maketitle

\begin{abstract}
Computational color constancy is a preprocessing step used in many camera systems. The main aim is to discount the effect of the illumination on the colors in the scene and restore the original colors of the objects. Recently, several deep learning-based approaches have been proposed to solve this problem and they often led to state-of-the-art performance in terms of average errors. However, for extreme samples, these methods fail and lead to high errors. In this paper, we address this limitation by proposing to aggregate different deep learning methods according to their output uncertainty. We estimate the relative uncertainty of each approach using Monte Carlo dropout and the final illumination estimate is obtained as the sum of the different model estimates weighted by the log-inverse of their corresponding uncertainties. The proposed framework leads to state-of-the-art performance on INTEL-TAU dataset.
\end{abstract}


%
\IEEEpeerreviewmaketitle

\section{Introduction}

The human eye can effectively adjust to changes in  visual conditions and different illuminations of the scene \cite{ebner2007color,alsam2010colour}. The well-acknowledged image processing problem of computational color constancy tries to mimic this ability in  digital cameras \cite{barnard1999practical}. The aim is to restore the `original' colors of the objects in a given scene and make it look like it was taken under a neutral white illumination. The main assumption which is usually made to reduce the complexity of the problem is that of one global uniform illumination across the whole scene \cite{barnard1999practical,gijsenij2011computational}.  Computational color constancy methods then can be decomposed into two steps: the first step is to estimate the global illumination of a given scene and the second step is simply to normalize the pixel values of the scene using the estimated illumination. As the second step is a straightforward transformation, the problem of color constancy can be reduced to estimating the illumination and, thus, is usually referred to as the illumination estimation problem \cite{barnard1999practical}. Over the past few years, different unsupervised \cite{choudhury2010color,qian2019finding,yang2015efficient,banic2017unsupervised,d4,d5,d7,d6,d8} and supervised \cite{46440,lou2015color,22,44,mine,f1,Barron2015ConvolutionalCC} techniques have been proposed to tackle this problem. 

Recently, with the advancement of data-driven methods and the availability of more labeled illumination estimation datasets \cite{nus,Hemrit2018RehabilitatingTC,laakom2019intel}, the direction of computational color constancy research has shifted towards learning-based approaches in general and deep learning-based methods in particular \cite{22,44,mine,f1,Barron2015ConvolutionalCC,qian2016deep,afifi2020deep}.  \acp{CNN} have been recently extensively used to estimate the global illumination of a given scene and have often outperformed  traditional statistical methods \cite{46440,44,f1}. While these methods usually yield low average error rates, their usage in industrial applications remains limited \cite{gao2017improving,afifi2019color}. This is due to the inability of a single model to accurately approximate the illuminations of all types of scenes and the failure of each of these approaches for some scene content \cite{mine,gao2017improving,afifi2019sensor}. 
\begin{figure}[t]
\centering
\includegraphics[width= 0.45\textwidth]{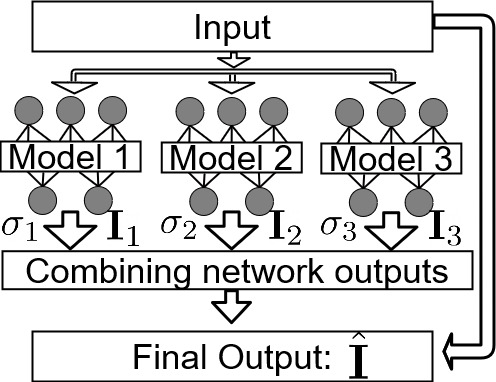}
\caption{Combining three different neural networks model outputs $\textbf{I}_1$, $\textbf{I}_2$, and $I_3$ using their predicted uncertainties $\sigma_1$, $\sigma_2$, and $\sigma_3$, respectively.}
\label{intro_fig}
\end{figure}

In this paper, we address this limitation by proposing a novel deep learning approach called Monte Carlo Dropout Ensemble\ac{MCDE}. In the proposed framework, we aggregate the estimates of different deep learning methods depending on the scene content and on the confidence of their predictions, as illustrated in Figure \ref{intro_fig}. To do so, we need to estimate the uncertainty of each model alongside its illumination estimation. To this end, we employ Monte Carlo dropout (MC-dropout) \cite{gal2015dropout,gal2016dropout} to obtain the output uncertainty of each \ac{CNN} model. Given both the illumination estimation and the model's uncertainty for each of the models, we propose a confidence score-based scheme to produce the final global illumination estimation. 

The use of MC-dropout for estimating the model's uncertainty does not require retraining of the available models. Thus, it can be used to model the uncertainty of any \ac{CNN}-based model trained with dropout \cite{gal2016dropout}. With the model confidence at hand, we can treat uncertain inputs and special cases explicitly. In this paper, we study as a  special case of our approach the combination of two state-of-the-art \ac{CNN}-based color constancy models, namely the \ac{BoCF} \cite{f1} and the \ac{FC4} \cite{44}, by weighting their relative illumination estimations according to their relative uncertainties. This yields a robust illumination estimation framework. 

The main contributions of this paper can be summarized as follows: 
\begin{itemize}
\item We propose a novel computational color constancy approach, called \acf{MCDE}, that aggregates different \ac{CNN}-based models based on their confidence in illumination estimation.
\item We propose using Monte Carlo dropout to approximate the uncertainty of different color constancy models.
\item To obtain the final estimate, we rely on the log-inverse of the uncertainty as a confidence score.
\item We study as a special case of our approach the combination of \acf{BoCF} and \acf{FC4}. The proposed approach yields state-of-the-art results on INTEL-TAU dataset.
\end{itemize}

The remainder of the paper is structured as follows. Section \ref{related} provides a brief review of \ac{CNN}-based color constancy approaches and Monte Carlo dropout.  Section \ref{proposed} presents the proposed approach \ac{MCDE}. Section \ref{experiments} presents the experimental setup and reports the evaluation results on INTEL-TAU dataset. Section \ref{conclusion} concludes the paper.

\section{Related work} \label{related}
\subsection{\ac{CNN}-based color constancy}

\acp{CNN} have been extensively used in the last decade to solve the computational color constancy problem \cite{22,44,mine,f1,Barron2015ConvolutionalCC}. The main task is to train a neural network that approximates the global illumination of the scene. \ac{CNN} are usually optimized in an end-to-end manner using back-propagation. We divide  \ac{CNN}-based approaches used to tackle the illumination estimation problem into two main categories: methods operating on local patches as input, called patch-wise methods  \cite{22,mine,bianco2017single,bianco2012color,bianco2014adaptive}, and methods operating directly on the full image as input, called single-pass methods \cite{lou2015color,44,f1}.

In the first group of methods, the models are trained on small patches of the scene to overcome the data shortage. Patch-based \acp{CNN} were first used for color constancy by Bianco et al. in \cite{22,bianco2017single}, where local estimates are generated from the small patches of the scene. The final estimate is pooled as the mean or median of these local estimates \cite{22} or passed to a support vector regressor \cite{bianco2017single} to estimate the illumination color. In \cite{mine}, an unsupervised pretraining phase is proposed to enhance the generalization of the model. A fully convolutional autoencoder with two branches, one for scene reconstruction and one for illumination estimation, are initially trained on the color constancy dataset augmented with ImageNet. Then, the illumination estimation branch is fine-tuned to approximate the patch-based illumination of the scene. This resulted in a better generalization capability and  a more robust model. Another patch-based \ac{CNN} approach is proposed by Bianco and Schettini \cite{bianco2012color,bianco2014adaptive}, which focuses specifically on face regions for color constancy.

In the second group of approaches, i.e., single-pass methods, networks are trained on the full image to approximate the illumination of the scene. These methods are faster as they require a single forward pass in the test phase. However, they usually operate on high resolution images and thus they have a higher number of parameters and higher memory requirements.  The early works in this category were proposed in \cite{lou2015color}, where a \ac{CNN} operating on the full image is trained to extract feature hierarchies to achieve robust color constancy. However, these methods rely on the conventional \ac{CNN} architecture, i.e., convolutional layers followed by max-pooling and a fully connected block in the end. Thus, in the inference phase, images need to be resized to predefined dimensions, which may introduce spatial distortions of image content and degrade the performance of the method. To overcome this limitation, recently \ac{FC4} \cite{44}  and \ac{BoCF} \cite{f1} were proposed. 

\ac{FC4} \cite{44} method relies on a fully convolutional topology using a confidence-weighted pooling layer. The network is trained end-to-end to incorporate learning the confidence of each patch of the image in a single pass and optimizing the weighted sum of the local estimates based on the confidences. 
In \cite{f1}, it is argued that the spatial information is not important in the color constancy context and, thus, it is discarded using a BoF layer \cite{bofp}. The resulting \ac{BoCF} network is composed of three blocks, namely the feature extraction, the Bag of Features, and the illumination estimation blocks. In the first block, a nonlinear transformation of the scene is produced. In the second block, a histogram representation of this transformation is compiled. This histogram is used in the third block to approximate the illumination. Moreover, an attention mechanism is used to adaptively learn to select the elements of the histogram which best encode the illuminant information.

Both \ac{FC4}  and \ac{BoCF} lead to a high performance in terms of average errors. However, for extreme samples, these methods fail and lead to high errors.  In this paper, we address this limitation by proposing a novel scheme to aggregate \ac{BoCF} and \ac{FC4} methods to deal with estimation ambiguities. We estimate the relative uncertainty of each approach using MC-dropout and the final illumination estimate is obtained as the sum of the different model estimates weighted by the log-inverse of their corresponding uncertainties. 

\subsection{Monte-Carlo dropout}
Dropout \cite{srivastava2014dropout,park2016analysis} is a regularization technique employed in many deep learning methods to avoid over-fitting and improve the robustness of the models.  Dropout is used in the training phase by discounting the output of some neurons in hidden layers with a fixed probability rate \cite{srivastava2014dropout}. This technique efficiently samples from an exponential number of different networks in a tractable and feasible way \cite{srivastava2014dropout}. 

In \cite{gal2016dropout}, it was shown that using dropout in \acp{CNN} can be interpreted as a Bayesian approximation of a Gaussian process \cite{bernardo2009bayesian,damianou2013deep,di2018gaussian}.  Applying dropout in the test time presents an approach to approximate the model uncertainty in deep learning without sacrificing neither the computational complexity nor the test accuracy.  For this approach, called MC-dropout, the inference phase is no longer deterministic as it depends on the randomly dropped-out neurons. For the same test example, the model yields different predictions each time. The variance or the standard deviation of the predictions can be interpreted as the uncertainty of the model. MC-dropout has been applied in many contexts, such as recurrent neural networks \cite{gal2016theoretically}, machine translation \cite{sennrich2016edinburgh}, and medical diagnostics \cite{yang2016fast,angermueller2015multi}.

In the color constancy context, both \ac{BoCF} and \ac{FC4} contain dropout layers and, thus, they can be extended with MC-dropout to estimate their confidence along with illumination output. With the model confidence at hand, we can treat uncertain inputs and special cases explicitly. However, it should be noted that using MC-dropout requires multiple forward passes of the model with each test input. This increases the time complexity of the algorithm. 

\section{Proposed framework} \label{proposed}
In the computational color constancy problem, the main task is to learn a function $f(\textbf{X})$, which takes an image $\textbf{X}$ as input and outputs an estimate of its illumination $\textbf{I}$. The \ac{CNN}-based approaches approximate the function $f(\textbf{X})$ with a neural network $f_\Theta(\textbf{X})$, where $\Theta$ is a set of parameters. The weights $\Theta$ are learned in an end-to-end manner using back-propagation.  \acp{CNN} have been recently extensively used in illumination estimation and have often led to state-of-the-art results and low average error rates. However, for extreme test samples, these methods fail and lead to high errors and to an undesired distortion of the colors of the image. This can be explained by the fact that a single model is unable to efficiently estimate the illumination for all types of scene content and all environments. 

In our framework, i.e., \ac{MCDE}, we tackle this limitation and propose combining different  \ac{CNN}-based color constancy models which have dropout layers by weighting their  relative illumination estimations according to their relative uncertainties. To estimate the uncertainty of each approach, we extend the original \ac{CNN} models using MC-dropout. Following this paradigm, we apply dropout in the test phase. When an unseen sample $\textbf{X}_1$ is presented,  $f_\Theta(\textbf{X}_1)$ is non-deterministic and yields  different outputs at each forward pass. For $\nu$ forward passes of $X_1 $, we have $\{\textbf{f}^1,\textbf{f}^2,..,\textbf{f}^{\nu}\}$ different estimates. The model illumination estimation can be computed as the average of the estimates:
\begin{equation}
   \textbf{I} = \frac{1}{\nu} \sum_{i=1}^{\nu} \textbf{f}^i.
\end{equation}
As $\textbf{I}$ is an RGB color and has three component $\textbf{I} = (\textbf{I}_r,\textbf{I}_g,\textbf{I}_b)$,  the standard deviation of each component can be computed as follows: 
\begin{equation}
   \sigma_m = \sqrt{ \frac{1}{\nu} \sum_{i=1}^{\nu} (\textbf{f}^i_m - \textbf{I}_m)^2 }, \hspace{2mm} for  \hspace{1mm} m=r,g,b.
\end{equation}
The standard deviation measures the amount of variation or dispersion of the set of predictions from the mean \cite{bland1996statistics}.  A confident model is expected to output similar values at each forward pass of $\textbf{X}_1$ and, thus, its corresponding standard deviations of  $\{\textbf{f}^1,\textbf{f}^2,..,\textbf{f}^{\nu}\}$ for each component are expected to be small, whereas if the input $\textbf{X}_1$ is ambiguous and the model is not confident, the predictions set  $\{\textbf{f}^1,\textbf{f}^2,..,\textbf{f}^{\nu}\}$ is expected to have high standard deviations for the components. Thus, the standard deviation can be used to model the component-wise uncertainty of $f_\Theta(\textbf{X})$ for the input $\textbf{X}_1$.  Moreover, the total uncertainty  $\mu$ of the model $f_\Theta(\textbf{X})$ on the prediction $\textbf{I}$ can be approximated as the product of the standard deviation along the three components, i.e., $\mu = \sigma_r\sigma_g\sigma_b $.

Let $ \{ \textbf{I}_1,...,\textbf{I}_K \}$  be the illumination predictions of K different \ac{CNN}-based models  and  $ \{ \sigma_1,...,\sigma_K \}$   their corresponding uncertainties. In \ac{MCDE}, we propose using a weighting scheme, where high weights are given to the more confident models and low weights are attributed to uncertain models. As a result, we decrease the dependency of the framework on a single model and provide a straightforward approach to leverage various models and combine them into a robust color constancy system.  Any function which is inversely proportional to the uncertainty can be used as a confidence score. The confidence scores of the K models can be computed as follows:
\begin{equation} \label{gg}
   c_i = g\left(\frac{1}{\sigma_i}\right), \hspace{2mm} for  \hspace{1mm} i=1,..,K
\end{equation}
where $g(.)$ can be a monotonic increasing function, e.g., linear or logarithm. So, the final estimate can be obtained as as the weighted sum of   $ \{ \textbf{I}_1,...,\textbf{I}_K \}$. The output illuminations of the models, $\textbf{I}_i$ for i=1,..,K  , have a norm of 1, i.e., $ ||\textbf{I}_i||_2=1$. To not distort the norm of the final prediction, we propose using the angular average. To this end, $ \{ \textbf{I}_i\}^K_{i=1}$ are converted to the spherical coordinate space $\{\textbf{Is}_i=(1,\phi_i, \varphi_i)\}^K_{i=1}$ via the following transformations:

\begin{equation}
   \phi_i = tan^{-1}\left( \frac{{\textbf{I}_i}_g}{{\textbf{I}_i}_r} \right), \hspace{2mm} for  \hspace{1mm} i=1,..,K
\end{equation}
\begin{equation}
   \varphi_i = tan^{-1}\left( \frac{ \sqrt {\textbf{I}_{ir} + \textbf{I}_{ig} } }{{\textbf{I}_i}_b}
   \right), \hspace{2mm} for  \hspace{1mm} i=1,..,K 
\end{equation}

The spherical coordinates $\textbf{Is}^{est}=(1,\phi, \varphi)$ of the illumination $\textbf{I}^{est}$ can be computed as follows:
\begin{equation}
   \phi = \sum_{i=1}^K c_i \phi_i. 
\end{equation}
\begin{equation}
   \varphi = \sum_{i=1}^K c_i \varphi_i. 
\end{equation}
The estimated illumination $\textbf{I}^{est}$ coordinates can thus be computed via the following inverse transformations: 
\begin{equation}
   \textbf{I}^{est}_r = sin(\varphi)cos(\phi).
\end{equation}
\begin{equation}
   \textbf{I}^{est}_g = sin(\varphi)sin(\phi).
\end{equation}
\begin{equation}
   \textbf{I}^{est}_b = cos(\varphi).
\end{equation}
$\textbf{I}^{est}$ is the output of \ac{MCDE} and is used to correct the colors of the input scene. The proposed framework enables us to leverage the benefits of the different models while dealing with estimation ambiguities. For each input, higher weights are given to the models with higher confidences and the contribution of the uncertain models are suppressed. This provides a simple yet effective way to adapt to the scene content and provide a robust approximation of the illumination.

\begin{figure}[]
\centering
\includegraphics[width=0.45\textwidth]{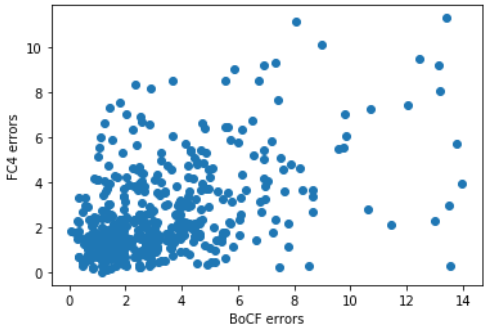}
\caption{Recovery angular errors of \ac{FC4} as a function of the Recovery angular errors of \ac{BoCF} on the first fold of INTEL-TAU dataset. }
\label{errors_fig}
\end{figure}
\begin{figure}[]
\centering
\includegraphics[width=0.45\textwidth]{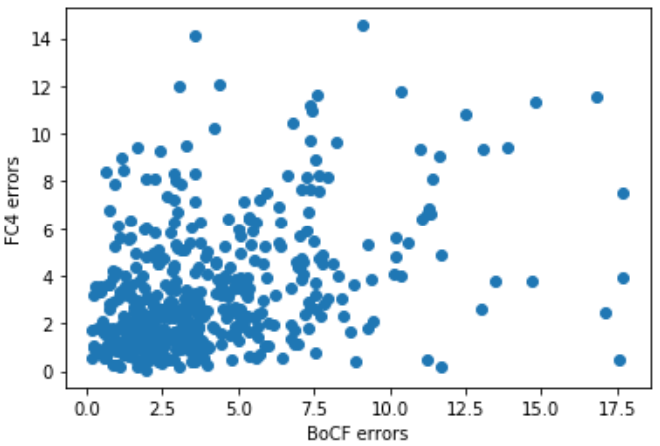}
\caption{Reproduction angular errors of \ac{FC4} as a function of the Reproduction angular errors of \ac{BoCF} on the first fold of INTEL-TAU dataset. }
\label{errors_fig1}
\end{figure}

\section{Experimental results and discussion} \label{experiments}
\subsection*{Dataset}
To evaluate the proposed approach, we use the INTEL-TAU dataset \cite{laakom2019intel}. It is the largest publicly available high-resolution illumination estimation dataset with 7022 images. Three different camera models were used, namely  Canon 5DSR, Nikon D810, and Sony IMX135. We evaluate the methods using the 10-folds non-random cross-validation protocol defined in \cite{laakom2019intel}.

\subsection*{Error metrics}
Similar to prior works, we report the mean of the best 25\%, the mean, the median, the trimean, and the mean of the worst 25\% of the Recovery angular error  $e_{recovery}$ \cite{21} between the ground truth illuminant and the estimated illuminant. For better insights on the performance of the methods, we also provide the same statistics with the recently proposed Reproduction angular error $e_{reproduction}$ \cite{finlayson2014reproduction}. The error metrics can be computed as follows:
\begin{equation}
       \text{$e_{recovery}$}(\textbf{I}^{gt},\textbf{I}^{est})= \cos^{-1} \left({ \frac{ \textbf{I}^{gt} \textbf{I}^{est}}{\| \textbf{I}^{gt} \| \|\textbf{I}^{est} \| } } \right) 
\end{equation}

\begin{equation}
     \text{$e_{reproduction}$}(\textbf{I}^{gt},\textbf{I}^{est})= \cos^{-1} \left({ \frac{ r(\textbf{I}^{gt} , \textbf{I}^{est})  \hspace{2mm} \textbf{n}}{\| r(\textbf{I}^{gt} , \textbf{I}^{est}) \| } } \right), 
\end{equation}
where $\textbf{I}^{gt}$ is the ground truth illumination, $\textbf{I}^{est}$ is the estimated illumination, $r(\textbf{I}^{gt} , \textbf{I}^{est}) = \textbf{I}^{gt}/ \textbf{I}^{est}$ is the element-wise division of $\textbf{I}^{gt}$ by $\textbf{I}^{est}$, and $\textbf{n}$ is the normalized unit vector, i.e., $\textbf{n} = [1/\sqrt3, 1/\sqrt3, 1/\sqrt3 ]^T$. 

\begin{figure*}[h]
\centering
\includegraphics[width=0.49\linewidth]{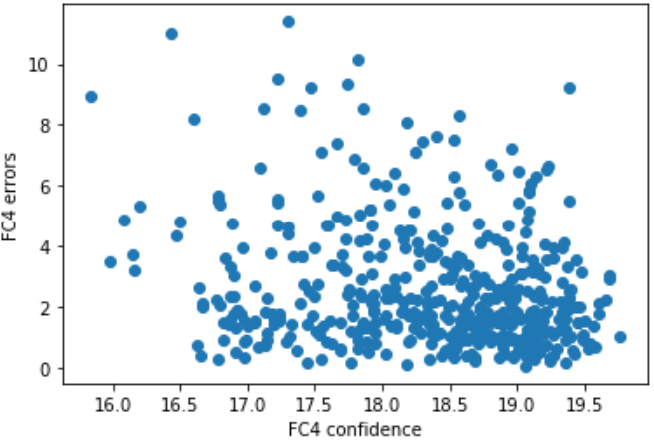}
\includegraphics[width=0.49\linewidth]{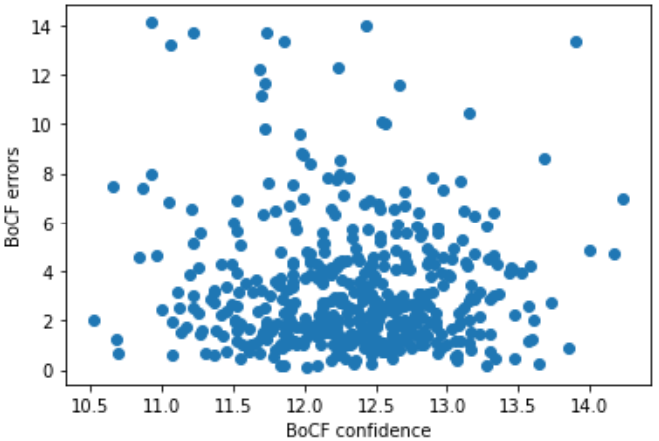}
\caption{The Recovery angular errors of \ac{FC4} and \ac{BoCF} as a function of the corresponding log-inverse confidence scores on the first fold of INTEL-TAU.}
\label{fig_1}
\end{figure*}

\begin{table*}[h]
	\caption{Results of benchmark methods on INTEL-TAU dataset using cross-validation protocol.}
		\label{tab:inteltauuncer}
 \centering	

\begin{tabular}{l|llllc||llllc}
                    & \multicolumn{5}{c||}{$e_{recovery}$}                                                       & \multicolumn{5}{c}{$e_{reproduction}$}                                                  \\ \hline
Method              & Best \newline 25\% & Mean & Med. & Tri. & W. \newline 25\% & Best \newline 25\% & Mean & Med. & Tri. & W. \newline 25\% \\
\hline
Grey-World  \cite{d4}        & 1.0                               & 4.9  & 3.9  & 4.1  & 10.5                            & 1.2                               & 6.1  & 4.9  & 5.2  & 13.0                            \\
White-Patch \cite{d5}     & 1.4                               & 9.4  & 9.1  & 9.2  & 17.6                            & 1.8                               & 10.0  & 9.5  & 9.8  & 19.2                            \\
Grey Edge    \cite{d7}    & 1.0  & 5.9  & 4.0  & 4.6  & 13.8& 1.2  & 6.8  & 4.9  & 5.5  & 13.5   \\
Grey Edge 2 \cite{d7} & 1.0  & 6.0  & 3.9  & 4.8  & 14.0 & 1.2   & 6.9  & 4.9  & 5.6  & 15.7 \\
Shades-of-Grey \cite{d6}  & 0.9 & 5.2  & 3.8  & 4.3  & 11.9                             & 1.1  & 6.3  & 4.7  & 5.1  & 13.9                            \\
Cheng et al. 2014 \cite{nus} & 0.7 & 4.5  & 3.2 & 3.5 & 10.6 &  0.9 & 5.5  & 4.0  & 4.4  & 12.7 \\
Weighted GE  \cite{d8} &  0.8                               & 6.1  & 3.7  & 4.6  & 15.1                            & 1.1                               & 6.9  & 4.5  & 5.4  & 16.5                            \\
Yang et al. 2015  \cite{yang2015efficient} &  0.6                               &  3.2  &  2.2  &  2.4  &  7.6                             &  0.7                               &  4.1  &  2.7  &  3.1  &  9.6                            \\
Color Tiger   \cite{banic2017unsupervised}        & 1.0     & 4.2    & 2.6  & 3.2  & 9.9    & 1.1      & 5.3  &  3.3  & 4.1  & 12.7   \\
Greyness Index  \cite{qian2019finding} &  0.5                               &  3.9    &  2.3  &  2.7  &  9.8                             &  0.6                               &  4.9  &  3.0  &  3.5  &  12.1   \\

PCC\_Q2 \cite{laakom2020probabilistic} &  0.6                               &  3.9    &  2.4  &  2.8  &  9.6                             &  0.7 &  5.1& 3.5  &  4.0  &  11.9  \\
\hline

FFCC\cite{46440} &  \textbf{0.4}   &  2.4    & \textbf{1.6}  & 1.8  & 5.6 &  \textbf{0.5}   & 3.0  & 2.1  & 2.3  & 7.1 \\

Bianco  \cite{22} &  0.9&  3.5   & 2.6  & 2.8  & 7.4                             &  1.1                            & 4.4 & 3.4  & 3.6  & 9.4 \\
C3AE \cite{mine} &  0.9&  3.4    & 2.7  & 2.8  & 7.0                             &  1.1& 3.9  & 3.3  & 3.5  & 8.8 \\

\ac{BoCF} \cite{f1} &  0.7&  2.4   & 1.9  & 2.0  & 5.1                           &  0.8    & 3.0& 2.3  &  2.5& 6.5 \\
\ac{FC4} (VGG16) \cite{44} &  0.6&  2.2    & 1.7  & 1.8  & 4.7  &  0.7 & 2.9  & 2.2  & 2.3  & 6.1 \\
\hline
\ac{MCDE} (linear)   &  0.6&  2.2    & 1.7  & 1.8  & 4.7   &  0.7  & 2.8  & 2.2  & 2.3  & 6.1 \\
\ac{MCDE} (log)  &  0.5&  \textbf{2.1}    & \textbf{1.6}  & \textbf{1.7}  & \textbf{4.5}  &  0.6  & \textbf{2.6}  & \textbf{2.0}  & \textbf{2.1}  & \textbf{5.4} \\
\end{tabular}
\end{table*}

\subsection*{Performance evaluation}
In our experiments, we evaluated the performance of the following state-of-the-art unsupervised approaches: Grey-World \cite{d4}, White-Patch \cite{d5}, Spatial domain \cite{nus}, Shades-of-Grey \cite{d6}, and Weighted Grey-Edge \cite{d8}, Greyness Index 2019  \cite{qian2019finding}, Color Tiger   \cite{banic2017unsupervised}, PCC\_Q2 \cite{laakom2020probabilistic}, and the method reported in \cite{yang2015efficient}.  In addition, we considered the learning-based approach  Fast Fourier Color Constancy  (FFCC) \cite{46440} and the four following \ac{CNN}-based approaches:  \ac{FC4} \cite{44}, Bianco \cite{22}, C3AE \cite{mine}, and BoCF \cite{f1}.

In our experiments, we consider the special case of \ac{MCDE} composed of K=2 models, namely \ac{FC4} and \ac{BoCF}. For the function $g$ defined in Eq \ref{gg}, we use the logarithm function, i.e., $g(x) = log(x)$. This variant of the framework is called \ac{MCDE} (log). In this variant, the confidence score are thus obtained as the log-inverse of the uncertainties. We also experimented with a basic variant of our framework, called \ac{MCDE} (linear), where $g$ is defined as the identity function. In this variant, the confidence scores $c_1$ and $c_2$ of \ac{FC4} and \ac{BoCF} defined in Eq. \ref{gg} are expressed directly as the inverse of the uncertainty, i.e., $c_1=\frac{1}{\sigma_1}$ and $c_2=\frac{1}{\sigma_2}$.  The number of forward passes per model, $\nu$, is fixed to 30 for both variants of our framework.


In the Figures \ref{errors_fig} and \ref{errors_fig1}, we plot the Recovery and the Reproduction angular errors $e_{recovery}$ and  $e_{reproduction}$  of \ac{FC4} as a function of the corresponding angular errors of \ac{BoCF} on the first fold of INTEL-TAU. As illustrated in the figure, for a large part of the test images \ac{FC4} and \ac{BoCF} have different extreme cases, i.e., they succeed and fail with different test images. This supports our assumption that by combining the methods in an optimal way, there is potential for improved results.  Combining models with different extreme cases enables \ac{MCDE} to deal with the different content and to output a robust estimation for more images.

Figure \ref{fig_1} illustrates the angular errors of \ac{FC4} and \ac{BoCF} as a function of the corresponding log-inverse confidences, ie., $g(x) = log(x)$. In the optimal case, the confidence would be high for samples with low errors and vice versa. We note that this is true in some cases and, especially for \ac{FC4}, the low errors tend to get high confidences. 
This supports our assumption that the log-inverse of the MC-dropout uncertainty used in \ac{MCDE} provides a useful tool to estimate the confidence of the color constancy models. Nevertheless, we see that the error-confidence correlation is not optimal and, in the future, other more advanced confidence measures should be developed and compared against the proposed MC uncertainty.  


\begin{figure*}[h]
\centering
\includegraphics[width= 0.9\textwidth]{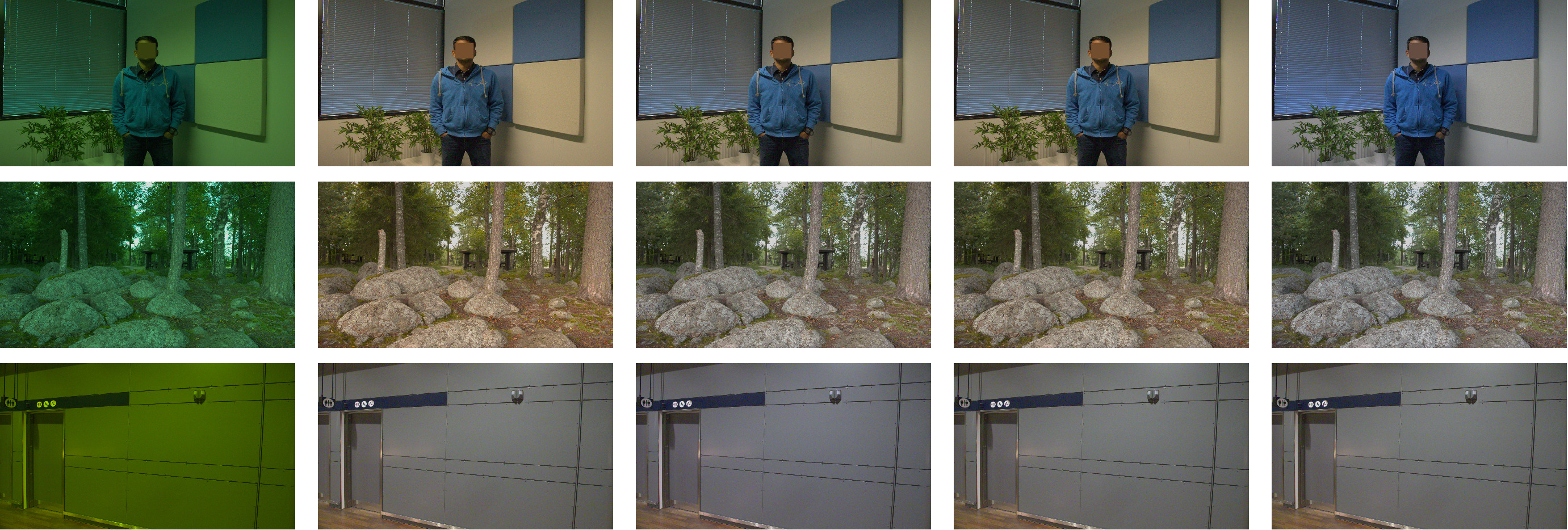}
\caption{Visual results on three sample of INTEL-TAU. From left to right: Input image, \ac{BoCF} output, \ac{FC4} output, \ac{MCDE} output, and ground truth image. Gamma correction was applied for visualization.}
\label{dd}
\end{figure*}
\begin{table*}[]
\setlength{\tabcolsep}{5pt}
\renewcommand{\arraystretch}{1}
	\caption{Results of CNN methods on INTEL-TAU dataset using cross-validation protocol.}
		\label{tab:worstsinario}
 \centering	
\begin{tabular}{l|lll||lll}
                    & \multicolumn{3}{c||}{$e_{recovery}$}                                                       & \multicolumn{3}{c}{$e_{reproduction}$}                                                  \\ \hline
Method & W. 25\% & W. 10\% & W. 5\%  & W. 25\% & W. 10\% & W. 5\% \\
\hline
FFCC\cite{46440}   & 5.6  & 7.9  & 9.8  &  7.1  & 10.2  & 12.4 \\
\ac{BoCF} \cite{f1}& 5.1  & 7.2  & 8.7  &  6.5  & 9.4  & 11.2 \\
\ac{FC4} \cite{44} & 4.7  & 6.4  & 7.8  &  6.1  & 8.3  & 10.1\\
\hline
\ac{MCDE} (linear)  & 4.7  & 6.4  & 7.8  &  6.1  & 8.3  & 9.6 \\
\ac{MCDE} (log) & 4.5  & 6.2  & 7.5  &  5.4  & 8.0  & 9.7 \\
\hline
Ideal & 3.8  & 5.3  & 6.6  &  4.7  & 6.7  & 8.2 \\

\end{tabular}
\end{table*}

Table \ref{tab:inteltauuncer} reports the competitive results of the state-of-the-art approaches along with the results of our proposed approaches on INTEL-TAU dataset. It can be seen that the competing \ac{CNN}-based approaches outperform other methods for both error functions except FFCC, which is a non-CNN learning-based method and it outperforms Bianco and C3AE especially in the mean of the worst 25\% metric. Our framework yields the state-of-the-art performance and the lowest error rates in all error metrics except the best 25\%, where FFCC achieves the best results. For handling the ambiguous input samples, the proposed framework outperforms the existing methods. This can be seen in terms of the mean of the worst 25\%.  In fact, the proposed framework achieves a $0.7^{\circ}$ improvement in this metric for the Reproduction error $e_{reproduction}$ compared to the prior methods. 

The proposed framework \ac{MCDE} aggregates \ac{FC4} and \ac{BoCF} via a weighting schema using their prediction uncertainty. Figure \ref{dd} presents a  visualization of the outputs of the images corrected with the different approaches. Compared to both of these methods, Table \ref{tab:inteltauuncer} shows that both variants of the proposed model achieve at least as low error rates as the best of these two. Thus, we see that relying on the MC-dropout uncertainty of the model indeed presents an effective way to model the confidence of the prediction. Moreover, by comparing both variants of our framework, we see that the log-inverse presents a more robust method for weighting the contribution of each model. In fact, this is illustrated especially in terms of the mean of the worst 25\%, where the log-inverse yields a $0.2^{\circ}$ improvement in the Recovery  error  $e_{recovery}$ and a  $0.7^{\circ}$ improvement in the Reproduction  error  $e_{reproduction}$.

To further highlight the robustness of the proposed framework on the extreme cases, we report in Table \ref{tab:worstsinario} the averages of  worst 25\%, worst 10\%, and worst 5\% for $e_{recovery}$ and $e_{reproduction}$. In addition, we report the results achieved by the  hypothetical ideal combination of \ac{FC4} and \ac{BoCF}, i.e., the results that would be obtained if always the better of the two methods was successfully selected. By comparing the errors rates of \ac{FC4}, \ac{BoCF}, and our approach, we note an improvement across all the metrics. \ac{MCDE} is indeed able to prune out some of the worst failure cases and combining different approaches using the proposed confidence scores improves the robustness of the approach. Moreover, we note that the ideal combination of \ac{FC4} and \ac{BoCF} achieves robust performance with more than $1^{\circ}$ improvement across all the metrics compared to each of the combined methods alone. This confirms that indeed both models succeed and fail in different contents justifies further the choice of combining these two models in our \ac{MCDE} framework. However, it should  also be noted that although \ac{MCDE} (log) yields an improved performance, there is still room for improvement by further improving the confidence estimate.

\section{Conclusion} \label{conclusion}
In this paper, we proposed a novel ensemble method for the computational color constancy problem called \ac{MCDE} aggregating different CNN models.  In the proposed framework,  we estimate the relative uncertainty of each model for a test sample using Monte Carlo dropout. This enables us to deal with estimation ambiguities.  The final illumination estimation is computed as the sum of the different models' estimates weighted by the log-inverse of their corresponding uncertainties. This yields  a robust illumination estimation system and reduces the dependency of the global framework using  a single model. Furthermore, the proposed approach addresses the limitation of prior works and the inability of a single model to generalize well for all types of scenes and can handle extreme cases. We evaluated the special case of \ac{MCDE} combining \ac{FC4} and \ac{BoCF} on the INTEL-TAU dataset. The proposed approach led to state-of-the-art performance. 

In future work, extensions of the proposed approach may include incorporating more CNN-based methods in \ac{MCDE}, dealing with the overconfidence bias, and exploring different alternatives to model the uncertainty of deep learning models in the color constancy context.

\section*{Acknowledgment} 
This  work  has been  supported  by  the  NSF-Business  Finland
Center for Visual and Decision Informatics (CVDI) project
AMALIA.

\bibliographystyle{IEEEtran}
\bibliography{citations}

\begin{thebibliography}{10}
\providecommand{\url}[1]{#1}
\csname url@samestyle\endcsname
\providecommand{\newblock}{\relax}
\providecommand{\bibinfo}[2]{#2}
\providecommand{\BIBentrySTDinterwordspacing}{\spaceskip=0pt\relax}
\providecommand{\BIBentryALTinterwordstretchfactor}{4}
\providecommand{\BIBentryALTinterwordspacing}{\spaceskip=\fontdimen2\font plus
\BIBentryALTinterwordstretchfactor\fontdimen3\font minus
  \fontdimen4\font\relax}
\providecommand{\BIBforeignlanguage}[2]{{%
\expandafter\ifx\csname l@#1\endcsname\relax
\typeout{** WARNING: IEEEtran.bst: No hyphenation pattern has been}%
\typeout{** loaded for the language `#1'. Using the pattern for}%
\typeout{** the default language instead.}%
\else
\language=\csname l@#1\endcsname
\fi
#2}}
\providecommand{\BIBdecl}{\relax}
\BIBdecl

\bibitem{ebner2007color}
M.~Ebner, \emph{Color constancy}.\hskip 1em plus 0.5em minus 0.4em\relax John
  Wiley \& Sons, 2007, vol.~7.

\bibitem{alsam2010colour}
A.~Alsam, ``Colour constant image sharpening,'' in \emph{2010 20th
  International Conference on Pattern Recognition}.\hskip 1em plus 0.5em minus
  0.4em\relax IEEE, 2010, pp. 4545--4548.

\bibitem{barnard1999practical}
K.~Barnard, \emph{Practical colour constancy}.\hskip 1em plus 0.5em minus
  0.4em\relax Simon Fraser University, 1999.

\bibitem{gijsenij2011computational}
A.~Gijsenij, T.~Gevers, and J.~Van De~Weijer, ``Computational color constancy:
  Survey and experiments,'' \emph{IEEE Transactions on Image Processing},
  vol.~20, no.~9, pp. 2475--2489, 2011.

\bibitem{choudhury2010color}
A.~Choudhury and G.~Medioni, ``Color constancy using standard deviation of
  color channels,'' in \emph{2010 20th International Conference on Pattern
  Recognition}.\hskip 1em plus 0.5em minus 0.4em\relax IEEE, 2010, pp.
  1722--1726.

\bibitem{qian2019finding}
Y.~Qian, J.-K. Kamarainen, J.~Nikkanen, and J.~Matas, ``On finding gray
  pixels,'' in \emph{Proceedings of the IEEE Conference on Computer Vision and
  Pattern Recognition}, 2019, pp. 8062--8070.

\bibitem{yang2015efficient}
K.-F. Yang, S.-B. Gao, and Y.-J. Li, ``Efficient illuminant estimation for
  color constancy using grey pixels,'' in \emph{Proceedings of the IEEE
  conference on computer vision and pattern recognition}, 2015, pp. 2254--2263.

\bibitem{banic2017unsupervised}
N.~Bani{\'c}, K.~Ko{\v{s}}{\v{c}}evi{\'c}, and S.~Lon{\v{c}}ari{\'c},
  ``Unsupervised learning for color constancy,'' \emph{arXiv preprint
  arXiv:1712.00436}, 2017.

\bibitem{d4}
J.~Cepeda-Negrete and R.~Sanchez-Yanez, ``Gray-world assumption on perceptual
  color spaces,'' in \emph{Image and Video Technology}, 2014, pp. 493--504.

\bibitem{d5}
A.~Rizzi, C.~Gatta, and D.~Marini, ``Color correction between gray world and
  white patch,'' in \emph{The International Society for Optical Engineering},
  2002.

\bibitem{d7}
J.~van~de Weijer, T.~Gevers, and A.~Gijsenij, ``Edge-based color constancy,''
  \emph{IEEE Transactions on Image Processing}, pp. 2207--2214, 2007.

\bibitem{d6}
G.~Finlayson and E.~Trezzi, ``Shades of gray and colour constancy,'' in
  \emph{Color Imaging Conference}, 2004, pp. 37--41.

\bibitem{d8}
A.~Gijsenij, T.~Gevers, and J.~Van De~Weijer, ``Physics-based edge evaluation
  for improved color constancy,'' in \emph{2009 IEEE Conference on Computer
  Vision and Pattern Recognition}.\hskip 1em plus 0.5em minus 0.4em\relax IEEE,
  2009, pp. 581--588.

\bibitem{46440}
J.~T. Barron and Y.-T. Tsai, ``Fast fourier color constancy,'' in \emph{IEEE
  Conference on Computer Vision and Pattern Recognition}, 2017.

\bibitem{lou2015color}
Z.~Lou, T.~Gevers, N.~Hu, M.~P. Lucassen \emph{et~al.}, ``Color constancy by
  deep learning.'' in \emph{BMVC}, 2015, pp. 76--1.

\bibitem{22}
S.~Bianco, C.~Cusano, and R.~Schettini, ``Color constancy using {CNN}s,'' in
  \emph{IEEE Conference on Computer Vision and Pattern Recognition Workshops},
  2015, pp. 81--89.

\bibitem{44}
Y.~Hu, B.~Wang, and S.~Lin, ``{FC}4: Fully convolutional color constancy with
  confidence-weighted pooling,'' in \emph{IEEE Conference on Computer Vision
  and Pattern Recognition}, 2017, pp. 4085 -- 4094.

\bibitem{mine}
F.~Laakom, J.~Raitoharju, A.~Iosifidis, J.~Nikkanen, and M.~Gabbouj, ``Color
  constancy convolutional autoencoder,'' in \emph{Symposium Series on
  Computational Intelligence}, 2019.

\bibitem{f1}
F.~Laakom, N.~Passalis, J.~Raitoharju, J.~Nikkanen, A.~Tefas, A.~Iosifidis, and
  M.~Gabbouj, ``Bag of color features for color constancy,'' \emph{IEEE
  Transactions on Image Processing, (Early Access) DOI:
  10.1109/TIP.2020.3004921}, 2020.

\bibitem{Barron2015ConvolutionalCC}
J.~T. Barron, ``Convolutional color constancy,'' \emph{IEEE International
  Conference on Computer Vision}, pp. 379--387, 2015.

\bibitem{nus}
D.~Cheng, D.~K. Prasad, and M.~S. Brown, ``Illuminant estimation for color
  constancy: why spatial-domain methods work and the role of the color
  distribution,'' \emph{Journal of the Optical Society of America. A, Optics,
  image science, and vision}, pp. 1049--1058, 2014.

\bibitem{Hemrit2018RehabilitatingTC}
G.~Hemrit, G.~Finlayson, A.~Gijsenij, P.~Gehler, S.~Bianco, B.~Funt, M.~Drew,
  and L.~Shi, ``Rehabilitating the colorchecker dataset for illuminant
  estimation,'' in \emph{Color and Imaging Conference}, 2018, pp. 350--353.

\bibitem{laakom2019intel}
F.~Laakom, J.~Raitoharju, A.~Iosifidis, J.~Nikkanen, and M.~Gabbouj,
  ``Intel-tau: A color constancy dataset,'' \emph{arXiv preprint
  arXiv:1910.10404}, 2019.

\bibitem{qian2016deep}
Y.~Qian, K.~Chen, J.-K. K{\"a}m{\"a}r{\"a}inen, J.~Nikkanen, and J.~Matas,
  ``Deep structured-output regression learning for computational color
  constancy,'' in \emph{2016 23rd International Conference on Pattern
  Recognition (ICPR)}.\hskip 1em plus 0.5em minus 0.4em\relax IEEE, 2016, pp.
  1899--1904.

\bibitem{afifi2020deep}
M.~Afifi and M.~S. Brown, ``Deep white-balance editing,'' in \emph{Proceedings
  of the IEEE/CVF Conference on Computer Vision and Pattern Recognition}, 2020,
  pp. 1397--1406.

\bibitem{gao2017improving}
S.-B. Gao, M.~Zhang, C.-Y. Li, and Y.-J. Li, ``Improving color constancy by
  discounting the variation of camera spectral sensitivity,'' \emph{JOSA A},
  vol.~34, no.~8, pp. 1448--1462, 2017.

\bibitem{afifi2019color}
M.~Afifi, B.~Price, S.~Cohen, and M.~S. Brown, ``When color constancy goes
  wrong: Correcting improperly white-balanced images,'' in \emph{Proceedings of
  the IEEE Conference on Computer Vision and Pattern Recognition}, 2019, pp.
  1535--1544.

\bibitem{afifi2019sensor}
M.~Afifi and M.~S. Brown, ``Sensor-independent illumination estimation for dnn
  models,'' \emph{arXiv preprint arXiv:1912.06888}, 2019.

\bibitem{gal2015dropout}
Y.~Gal and Z.~Ghahramani, ``Dropout as a bayesian approximation: Insights and
  applications,'' in \emph{Deep Learning Workshop, ICML}, vol.~1, 2015, p.~2.

\bibitem{gal2016dropout}
------, ``Dropout as a bayesian approximation: Representing model uncertainty
  in deep learning,'' in \emph{international conference on machine learning},
  2016, pp. 1050--1059.

\bibitem{bianco2017single}
S.~Bianco, C.~Cusano, and R.~Schettini, ``Single and multiple illuminant
  estimation using convolutional neural networks,'' \emph{IEEE Transactions on
  Image Processing}, vol.~26, no.~9, pp. 4347--4362, 2017.

\bibitem{bianco2012color}
S.~Bianco and R.~Schettini, ``Color constancy using faces,'' in \emph{2012 IEEE
  Conference on Computer Vision and Pattern Recognition}.\hskip 1em plus 0.5em
  minus 0.4em\relax IEEE, 2012, pp. 65--72.

\bibitem{bianco2014adaptive}
------, ``Adaptive color constancy using faces,'' \emph{IEEE transactions on
  pattern analysis and machine intelligence}, vol.~36, no.~8, pp. 1505--1518,
  2014.

\bibitem{bofp}
N.~Passalis and A.~Tefas, ``Learning bag-of-features pooling for deep
  convolutional neural networks,'' in \emph{IEEE International Conference on
  Computer Vision}, 2017.

\bibitem{srivastava2014dropout}
N.~Srivastava, G.~Hinton, A.~Krizhevsky, I.~Sutskever, and R.~Salakhutdinov,
  ``Dropout: a simple way to prevent neural networks from overfitting,''
  \emph{The journal of machine learning research}, vol.~15, no.~1, pp.
  1929--1958, 2014.

\bibitem{park2016analysis}
S.~Park and N.~Kwak, ``Analysis on the dropout effect in convolutional neural
  networks,'' in \emph{Asian conference on computer vision}.\hskip 1em plus
  0.5em minus 0.4em\relax Springer, 2016, pp. 189--204.

\bibitem{bernardo2009bayesian}
J.~M. Bernardo and A.~F. Smith, \emph{Bayesian theory}.\hskip 1em plus 0.5em
  minus 0.4em\relax John Wiley \& Sons, 2009, vol. 405.

\bibitem{damianou2013deep}
A.~Damianou and N.~Lawrence, ``Deep gaussian processes,'' in \emph{Artificial
  Intelligence and Statistics}, 2013, pp. 207--215.

\bibitem{di2018gaussian}
A.~Di~Martino, E.~Bodin, C.~H. Ek, and N.~D. Campbell, ``Gaussian process deep
  belief networks: A smooth generative model of shape with uncertainty
  propagation,'' in \emph{Asian Conference on Computer Vision}.\hskip 1em plus
  0.5em minus 0.4em\relax Springer, 2018, pp. 3--20.

\bibitem{gal2016theoretically}
Y.~Gal and Z.~Ghahramani, ``A theoretically grounded application of dropout in
  recurrent neural networks,'' in \emph{Advances in neural information
  processing systems}, 2016, pp. 1019--1027.

\bibitem{sennrich2016edinburgh}
R.~Sennrich, B.~Haddow, and A.~Birch, ``Edinburgh neural machine translation
  systems for wmt 16,'' \emph{arXiv preprint arXiv:1606.02891}, 2016.

\bibitem{yang2016fast}
X.~Yang, R.~Kwitt, and M.~Niethammer, ``Fast predictive image registration,''
  in \emph{Deep Learning and Data Labeling for Medical Applications}.\hskip 1em
  plus 0.5em minus 0.4em\relax Springer, 2016, pp. 48--57.

\bibitem{angermueller2015multi}
C.~Angermueller and O.~Stegle, ``Multi-task deep neural network to predict cpg
  methylation profiles from low-coverage sequencing data,'' in \emph{NIPS MLCB
  workshop}, 2015.

\bibitem{bland1996statistics}
J.~M. Bland and D.~G. Altman, ``Statistics notes: measurement error,''
  \emph{Bmj}, vol. 312, no. 7047, p. 1654, 1996.

\bibitem{21}
S.~Hordley and G.~Finlayson, ``Re-evaluating colour constancy algorithms,'' in
  \emph{International Conference on Pattern Recognition}, 2004.

\bibitem{finlayson2014reproduction}
G.~Finlayson and R.~Zakizadeh, ``Reproduction angular error: An improved
  performance metric for illuminant estimation,'' \emph{perception}, 2014.

\bibitem{laakom2020probabilistic}
F.~Laakom, J.~Raitoharju, A.~Iosifidis, U.~Tuna, J.~Nikkanen, and M.~Gabbouj,
  ``Probabilistic color constancy,'' \emph{arXiv preprint arXiv:2005.02730},
  2020.

\end{thebibliography}

\end{document}